\documentclass[10pt,a4paper]{article}

\usepackage{cogsci}

\usepackage{array,multirow}
\usepackage{apacite}
\usepackage{graphicx}
\usepackage{lineno,hyperref}
\usepackage{amsmath} 
\usepackage{natbib}
\usepackage{eurosym}
\usepackage{graphicx}
\usepackage{wrapfig}
\usepackage[english]{babel}
\usepackage[center]{caption}
\usepackage{mathtools}
\usepackage{mathptmx}
\usepackage{latexsym}
\usepackage{url}
\usepackage{multirow}
\usepackage{amssymb}
\usepackage{capt-of, blindtext}
\usepackage{pdfpages}
\usepackage{setspace}
\usepackage{lineno}
\usepackage{algorithmic}
\usepackage{algorithm}
\usepackage{fullpage}
\usepackage{kpfonts}
\usepackage{setspace}
\usepackage{algorithmic}
\usepackage{algorithm}
\usepackage{bbold}

\usepackage{microtype}

\definecolor{brickred}{rgb}{0.8, 0.25, 0.33}

\title{Order Effects in Bayesian Updates}
 
\author{
        {\large \bf Catarina Moreira$^1$ (catarina.pintomoreira@qut.edu.au)} \AND 
        {\large \bf Jose Acacio de Barros$^2$ (debarros@sfsu.edu)}  \\
        $^1$School of Information Systems, Queensland University of Technology, Brisbane, Australia\\
        $^2$School of Humanities and Liberal Studies, San Francisco State University, San Francisco, CA, USA \\ \\
{\large \bf Paper accepted at the } \\
 {\textit Proceedings of the 43rd Annual Meeting of the Cognitive Science Society, 2021} \\
 {Link: \url{https://escholarship.org/uc/item/9cc2v2xh}}   
}

\begin{document}

\maketitle 

\begin{abstract} % MAX 150 words

Order effects occur when judgments about a hypothesis's probability given a sequence of information do not equal the probability of the same hypothesis when the information is reversed. Different experiments have been performed in the literature that supports evidence of order effects.

We proposed a Bayesian update model for order effects where each question can be thought of as a mini-experiment where the respondents reflect on their beliefs. We showed that order effects appear, and they have a simple cognitive explanation: the respondent's prior belief that two questions are correlated.

The proposed Bayesian model allows us to make several predictions: (1) we found certain conditions on the priors that limit the existence of order effects; (2) we show that, for our model, the QQ equality is not necessarily satisfied (due to symmetry assumptions); and (3) the proposed Bayesian model has the advantage of possessing fewer parameters than its quantum counterpart.

\textbf{Keywords: order effects; quantum cognition; Bayesian networks; Bayesian updates} 

\end{abstract}

\section{Introduction }

The application of Quantum Mechanic's mathematical principles in areas outside of physics has been getting increasing attention in the scientific community in a field called Quantum Cognition~\citep{Busemeyer12book,Moreira20NN,Moreira20cogsci}. These principles have been applied to explain paradoxical situations that cannot be easily explained through classical probability theory \citep{Moreira17faces,Moreira15}.  Quantum principles have also been adopted in many different domains, such as Cognitive Psychology~\citep{busemeyer06,Busemeyer09,Pothos13}, Economics~\citep{Khrennikov09eco,Haven13}, Biology~\citep{Asano12model,Asano15}, and Information Retrieval~\cite{Bruza09quantum,Bruza13}, to name a few. 

\subsection{Order Effects}

One of those paradoxical situations is {\it order effects}.
Order effects occur when judgments about a hypothesis's probability given a sequence of information do not equal the probability of the same hypothesis when the given information is reversed.
%Order effect is the change to the answer to a question $Q_1$ if it is asked by itself or following another question $Q_2$. In other words, given two questions $Q_1$ and $Q_2$, the order we ask them alters their answer. 

In purely classical models, such as the one examined by \cite{trueblood2011quantum}, this poses a problem. Since classical probability theory is based on set theory, this means that it is commutative. That is, for some question $Q3$ and two questions $Q_1$ and $Q_2$: $P\left(~Q_1 \cap Q_2~|~Q3~\right) = P\left(~Q_2 \cap Q_1~|~Q3~\right)$ since $P( Q_1 \cap Q_2) = P( Q_2 \cap Q_1$). This commutativity poses a challenge to model order effects, because in order to have $P\left(~Q3~|~Q_1 \cap Q_2~\right) = P\left(~Q3~|~Q_2 \cap Q_1~\right)$, then using Bayes Rule, one would need to satisfy the following relationship~\citep{trueblood2011quantum}:
\begin{equation}
     P( Q3 | ~Q_1 \cap Q_2~) =  P( Q3 |  ~Q_2 \cap Q_1 ),
\end{equation}
which implies that
\begin{equation}
   \frac{ P( Q_1 \cap Q_2 | Q3 ) P( Q3 ) }{P( Q_1 \cap Q_2 )} = \frac{ P(  Q_2 \cap Q_1 | Q3 ) P(Q3) }{P(Q_2 \cap Q_1)}.
\end{equation}

To accommodate these paradoxical findings, some researchers turned to non-standard probability theories. One such example is quantum probability, a theory based on operators' measures in Hilbert spaces instead of sample sets. Quantum probability models provide many advantages over their classical counterparts~\citep{Busemeyer15comparison,Moreira18}. They can represent events in vector spaces through a superposition state, which comprises all events at the same time. In quantum mechanics, the superposition principle refers to the property that particles must be in an indefinite state: a particle can be in different states at the same time. 

From a psychological perspective, a quantum superposition is interpreted as related to the feeling of confusion, uncertainty, or ambiguity~\citep{Busemeyer12book}. The vector space representation does not obey the distributive axiom of Boolean logic and to the law of total probability. Instead, events satisfy a quantum lattice structure and not a Boolean algebra. This vector representation enables the construction of more general models that can mathematically explain cognitive phenomena such as order effects~\citep{Busemeyer09,Khrennikov09sure}.

\subsection{Quantum Probability Models to Accommodate Order Effects}

One of the quantum approaches used to explain order effects is the quantum projection model~\citep{Pothos13}. In this approach, we start with a Hilbert space, $\mathcal{H}$, whose dimensions correspond to the number of possible responses to a question (e.g., $\mathbb{C}^2$). A Hermitian operator (called an observable operator) models each question in $\mathcal{H}$, and if two questions are to exhibit order effect, those operators are non-commuting. Because of the spectral theorem, such operators can be written as linear combinations of projection operators, with the eigenvalues multiplying each projector. Thus, in the quantum model for order effects, when an initial state, denoted by a vector (or density operator) in the Hilbert space, is subject to a question, the model treats it as a measurement represented by the operator. After a measurement, the initial state collapses into one of the subspaces of the observable operator. This new collapsed state is used to compute the probabilities of the second operator. Since projections are not unitary and do not commute, measuring one observable before another changes the measurements' outcomes. The non-unitarity of non-commuting quantum measurements leads, in essence, to order effects. We refer the reader to \cite{Pothos13} for details about the quantum model. 

\subsection{Contributions}

A question remains as to whether one needs quantum models to describe order effects~\citep{Moreira17order}. Can one use classical probability theory without constructing contrived models? Some previous works attempt to answer this question. In \citet{Costello18}, the authors argue that order effects result from classical probability theory with some noise, and in \citet{Kellen20}, the authors discuss a class of repeat-choice models that can provide approximate results to the quantum models. This paper tries to answer this question positively. We do so by constructing a Bayesian update model that accounts for order effects. The proposed model allows us to make several predictions: certain conditions on the priors limit the existence of order effects, and the proposed Bayesian model has fewer parameters than its quantum counterparts. 

\subsection{Paper Outline}

This paper is organized as follows. In the next section, we use Bayesian Networks to show a fallacy in the argument that order effects cannot be derived from classical probability theory. This Bayesian Network representation leads to the intuition that Bayesian probability updates can account for order effects. In the following section, we construct a Bayesian update model that offers further insights into the cognitive origins of order effects and show a series of predictions from the model. 
Finally, we end this article by summarizing the most relevant contributions presented throughout this study.

\section{Fallacies in Order Effects} \label{sec:BNs}

Consider the following experimental scenario. A set of participants is required to answer three questions, $Q_0$, $Q_1$, and $Q_2$ consecutively. One set of participants is presented with the questions $Q_0 \rightarrow Q_1 \rightarrow Q_2$, while the other participants are presented with the same questions, but in reversed order, $Q_0 \rightarrow Q_2 \rightarrow Q_1$. Some researchers argue the above experimental contexts should attain similar outcomes for questions $Q_1$ and $Q_2$ since, in classical probability theory, the joint distribution between two events is commutative. In other words, it is argued that the expected probability outcomes should be $P( Q_1, Q_2 | Q_0) = P( Q_2, Q_1 | Q_0)$. We claim that this premise is a fallacy due to the following reasons:
%In other words, in such experimental contexts, researchers argue that the expected probability outcomes should be $P( Q_1, Q_2 | Q_0) = P( Q_2, Q_1 | Q_0)$. We claim that this premise is a fallacy due to the following reasons:

\vspace{-0.1cm}
\begin{itemize}
    \item \textbf{There is no causality in probability theory.} Probability theory describes the likelihood of outcomes of random phenomena. Since events are represented as sets, there is no notion of temporality (order) between events, and, consequently, probability theory, by itself, says nothing about causality. The joint $Q_1 \cap Q_2$ means that both events occur, without any implications to time or cause \& effect (order) between the events.
    
    \item \textbf{There is no temporality in probability theory.} Order effects experiments imply the notion of time: one question needs to be asked after another. Again probability theory, due to its set-based foundation, nothing says about time. Conditional probability is simply a measure under additional information. It implies symmetry, and for that reason, it does not imply temporal order. Temporality is part of a model, and not part of probability theory itself. 
\end{itemize}

A good set of tools to visualize these fallacies are Bayesian networks (BNs), which are models that represent a compact full joint probability distribution through a directed acyclic graphical structure. Its underlying mathematical principles are Bayes rule and conditional independence. 

\begin{figure}[!h]
\centering
%\resizebox{\columnwidth}{!} {
    \includegraphics[scale=0.45]{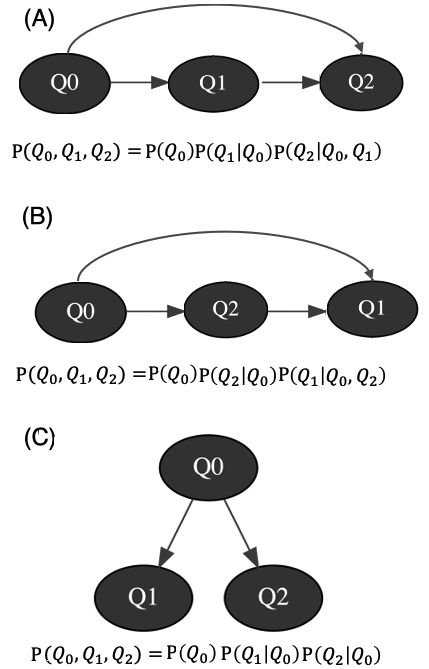}
%    }
    \caption{Different representations of a joint probability distribution portraying different assumptions between random variables. }
    \label{fig:bns}
\end{figure}

%Another fallacy that is often applied is related to the meaning of the full joint probability distribution, in this case $P( Q_0, Q_1, Q_2)$.

Given different full joint probability distributions, $P( Q_0, Q_1, Q_2)$, each distribution will be associated to a specific configuration of random variables. Consider the Bayesian networks in Figure~\ref{fig:bns}. They all represent the full joint probability distribution $P(Q_0, Q_1, Q_2)$, but with different underlying probabilistic graphical models. It is not correct to say that the full joint probability distribution in  Figure~\ref{fig:bns}-(A) has to be same as in  Figure~\ref{fig:bns}-(B). First, because they represent different experimental scenarios. For that reason, they yield different measures over the sample spaces. Second, the priors affecting the random variables change in both scenarios. It is a fallacy to assume that these two models should represent the same experimental context.

For example, the equation expressing the full joint of the network in Figure~\ref{fig:bns}-(A) is given by $P( Q_0, Q_1, Q_2 ) = P(Q_0)P(Q_1|Q_0)P(Q_2|Q_0,Q_1)$. In the same way, the equation representing the network in Figure~\ref{fig:bns}-(B) is given by  $P'( Q_0, Q_1, Q_2 ) = P'(Q_0)P'(Q_2|Q_0)P'(Q_1|Q_0,Q_2)$. Thus, it is a fallacy to assume that $P( Q_0, Q_1, Q_2 ) = P'( Q_0, Q_1, Q_2 ) $.

%\begin{equation}
% P( Q_0, Q_1, Q_2 ) = P(Q_0)P(Q_1|Q_0)P(Q_2|Q_0,Q_1),
% \end{equation}
% and the equation  of network in Figure~\ref{fig:bns}-(B) is given by
% \begin{equation}
%        P'( Q_0, Q_1, Q_2 ) = P'(Q_0)P'(Q_2|Q_0)P'(Q_1|Q_0,Q_2).
%\end{equation}
% It is a fallacy to assume that $P( Q_0, Q_1, Q_2 ) = P'( Q_0, Q_1, Q_2 ) $.

In this paper, we argue that to model order effect between three questions, $Q_0$, $Q_1$, and $Q_2$, we need to have a causal reasoning network structure as depicted in Figure~\ref{fig:bns}-(C). This means that order effects can be observed when:
\begin{itemize}
    \item $Q_0$ is correlated with $Q_1$ and $Q_2$, and
    \vspace{-0.1cm}
    \item $Q_1$ and $Q_2$ are conditionally independent, given $Q_0$, $(Q_1 \Perp Q_2 | Q_0)$.
\end{itemize}
These two conditions imply that, if there are \textit{no order effects}, then the following equation should hold:

\vspace{-0.5cm}
\begin{equation}
    \begin{split}
        P(Q_2 | Q_0 ) = P( Q_2 | Q_0, Q_1 ) \\
        P(Q_1 | Q_0 ) = P( Q_1 | Q_0, Q_2 )
    \end{split}
    \label{eq:order_effects}
\end{equation}

If Equation~\ref{eq:order_effects} is violated, then it suggests some correlation between $Q_1$ and $Q_2$, which may be derived from the degree of uncertainty that the decision-maker is experiencing. In the next section, we will use these notions to formulate a Bayesian update model to accommodate order effects.

\section{Order Effects in Bayesian Updates}\label{sec:updates}

\begin{table*}[!h]
\resizebox{2\columnwidth}{!} {
\begin{tabular}{l|c|c|c l|c|c|}
\cline{2-3} \cline{6-7}
 & \multicolumn{2}{c|}{\textbf{
                    \begin{tabular}[c]{@{}c@{}}
                        Scenario 1:\\ 
                        Clinton (C) $\rightarrow$ Gore (G) 
                    \end{tabular}}} 
&  &  & \multicolumn{2}{c|}{\textbf{
                    \begin{tabular}[c]{@{}c@{}}
                        Scenario 2:\\ 
                        Gore (G) $\rightarrow$ Clinton (C)
                    \end{tabular}}} \\ 
 \cline{2-3} \cline{6-7} 
 
 & \textbf{P( G = yes $|$ C )} & \textbf{P( G = no $|$ C )} & ~~~~~ & ~~~~~  & \textbf{P( C = yes $|$ G )} & \textbf{P( C = no $|$ G )} \\ \cline{1-3} \cline{5-7} 
 
    \multicolumn{1}{|l|}{\textbf{P( C = yes )}} & 0.4899 & 0.0447 & \multicolumn{1}{l|}{} & \textbf{P( G = yes )} & 0.5625 & 0.1991 \\ \cline{1-3} \cline{5-7} 
\multicolumn{1}{|l|}{\textbf{P( C = no )}} & 0.1767 & 0.2886 & \multicolumn{1}{l|}{} & \textbf{P( G = no )} & 0.0255 & 0.2130 \\ \cline{1-3} \cline{5-7} 
\end{tabular}
}
\caption{Order-effect experiments reported in the work of \citet{Moore02}. The random variable \textit{Clinton} represents the question \textit{Is [former US president Bill] Clinton honest and trustworthy?}; and \textit{Gore} represents the question \textit{Is [former US vice-president Al] Gore honest and trustworthy?}. A total of 501 participants answered the questions in each scenario, making a total of 1002 participants in the experiments.}
\label{tab:order_effects_lit}
\end{table*}

To understand the main idea, let us think of the well-known Clinton/Gore order-effect example, which corresponds to an order effect experiment originally conducted by~\citet{Moore02}. Consider the following two questions that were used in the experiment:
\begin{description}
\item[$Q_1$:] Is [former US president Bill] Clinton honest and trustworthy? 
\item[$Q_2$:] Is [former US vice-president Al] Gore honest and trustworthy? 
\end{description}
Experimental data shows that we get disparate results if we ask the above questions in different orders (Table~\ref{tab:order_effects_lit}) \cite{Wang14}. Namely, if we ask $Q_1$ before $Q_2$, the probability that $Q_1$ is answered in the positive is $53\%$. However, if we ask $Q_1$ after $Q_2$, i.e., if we change the order, $Q_1$ is answered positively $59\%$ of the time. Similarly, $Q_2$ gets $75\%$ if asked first and $65\%$ if asked after $Q_1$. The change in probabilities of $Q_1$ and $Q_2$ with the order they are answered is known as the order effect. 

How can we understand the order effect from a Bayesian point of view? A simple Bayesian model, such as the one presented by \cite{trueblood2011quantum},  cannot account for order effects. As discussed previously, the reason is simple: from a set-theoretic point of view, the conjunction $A\cap B$ is the same as $B \cap A$, so there is no order in it. 

So, does that mean that it is impossible to obtain order effects with Bayesian probabilities? The answer is no. To see this, consider the following circumstance. Imagine we ask a participant a question they are not sure of its answer. After some reflection, they may answer yes to it. However, this does not mean they are sure of their answer. For example, if after a moment of pondering, they answer ``yes'' to the question, they weighed the evidence for and against it and found a preponderance toward yes. Thus, we can think of the process of answering a question as a reflection on our beliefs. It is a type of mini-experiment. As such, after this mini-experiment, their beliefs may change. Consequently, they need to update their prior. Being asked and answering a question updates the prior to a posterior. 

To implement the above idea to produce an order effect, we follow the belief update described in \cite{morris1974decision,morris1977combining,Barros13}. Imagine a participant being asked the two questions, $Q_1$ and $Q_2$. Before answering $Q_1$, the participant has an unconscious prior joint probability $P(Q_1, Q_2)$. Once they reflect on $Q_1$, their prior is updated to a posterior $P'(Q_1, Q_2)$. This new posterior is the one used to compute the answer to $Q_2$. If the prior joint probability has $Q_1$ and $Q_2$ correlated (as is likely the case of Gore and Clinton), then the expectation for $Q_2$ will likely be different. The same would happen if $Q_2$ were asked first, with a new posterior  $P''(Q_1, Q_2)$ giving rise to a different expectation for $Q_1$. 

Let us describe mathematically the above intuition. We start with a participant, Alice, who will be asked to decide between the two questions, $Q_1$ and $Q_2$. Alice starts the experiment with a prior joint probability distribution $P(Q_1=x;Q_2=y|\kappa)$ conditioned on Alice's current knowledge $\kappa$, where $Q_1$ and $Q_2$ are $0/1$-valued random variables representing Alice's two possible answers, ``yes'' ($1$) and ``no'' ($0$). We will denote $P(Q_1=y)$ as $P(q_1)$ and the complementary probability, $P(Q_1=n)$ as $P(\overline{q}_1)$. Prior to being asked a question, Alice is unaware of her belief for each answer, encoded by the prior. It is not until she is asked that, by thinking about her belief, Alice comes up with one of two possible answers and perhaps a measure of her belief. Thus, Alice's choice of ``yes'' or ``no'' and her reflection requires a belief update. According to Bayes's theorem, Alice's prior became the posterior $P'(Q_1=x,Q_2=y|Q_1=z)$. Assuming a simple linear likelihood function, the updated posterior for the order $Q_1$ and then $Q_2$ becomes
\begin{equation}
    P'_{O1}(x,y|z)= \frac{P(Q_1=z)P(z,y)}
    {\sum_{x',y'} P(Q_1=x') P(Q_1=x',Q_2=y')},
\end{equation}
where we are using the simplifying notation $P'(Q_1=x,Q_2=y|Q_1=z)=P'(x,y|z)$ when there is no ambiguity in meaning, and the subscript $O_1$ for the order $Q_1$ first. Similarly, for the order $Q_1 \rightarrow Q_2$  we get
\begin{equation}
    P'_{O2}(x,y|z)= \frac{P(Q_2=z)P(z,y)}
    {\sum_{x',y'} P(Q_2=x') P(Q_1=x',Q_2=y')},
\end{equation}
for the inverse order, with $Q_2$ first. 

For simplicity, let us assume that the prior has well the following expectations.
\begin{align}
 P(q_1) = P(q_1,q_2) + P(q_1, \overline{q}_2) & = a, \\
 P(q_2) = P(\overline{q}_1,q_2) +P(\overline{q}_1,q_2) & = b, \\
 P(q_1,q_2) & = c.
\end{align}
$a$ and $b$ are the probabilities of $Q_1$ and $Q_2$ as true when asked first, respectively, and $c$ is related to their correlation, as it equals their joint moment. For instance, $c=1$ if both $Q_1$ and $Q_2$ are true. 

From Eq. 6-8, one can compute the remaining values of the full joint distribution by solving the following linear equation system:
\begin{equation}
\begin{cases} 
   P(q_1, q_2) + P(q_1,\overline{q}_2) +  P(\overline{q}_1,q_2) +  P(\overline{q}_1,\overline{q}_2) = 1  \\ 
    P(q_1) = P(q_1,q_2) +  P(q_1,\overline{q_2}) = a\\ 
    P(q_2) = P(q_1,q_2) +  P(\overline{q}_1,q_2) = b \\ 
    P(q_1,q_2) =c \\
\end{cases}
\label{eq:linear_solver}
\end{equation}

From Eq.~\ref{eq:linear_solver}, we can find that $P(q_1, q_2) = c$, $P(q_1, \overline{q}_2) = a - c$, $P(\overline{q}_1, q_2) = b - c$, and $ P(\overline{q}_1, \overline{q}_2) = 1 - a - b + c$. By substituting the values of Eq. 4 by the outputs of our linear solver, then we obtain Eq.~\ref{eq:order_effectQ2} from which it is straightforward to show that there is an order effect. For instance, for $Q_1$ first, the probability of a $true$ answer is $a$, but if one asks $Q_2$, after the update, the probability of being $true$ is
\begin{equation}
    P'(q_2|Q_1) =  \frac{(2a-1)c+(1-a)b}{2 a^2-2a+1}.
    \label{eq:order_effectQ2}
\end{equation}

%For $P'(q_1| Q_2)$, we obtain
%\begin{equation}
%    P'(q_1|Q_2) = \frac{a^2}{2 a^2-2a+1}.
%\end{equation}
We can see from the above equations that the updated probability for $Q_2$, $P'(q_2|Q_1)$, is not the same as it was before, $P(q_2)$. What are the conditions for the probability to be the same, i.e., for not having an order effect? 
\textit{No order effect} can be verified if the following conditions are satisfied:

\begin{itemize}
    \item Both questions are statistically independent, $Q_1 
    \Perp Q_2$, and the following conditions are satisfied:
    \begin{itemize}
        \item $P'(q_2|Q_1) = P(q_2)$, the belief of answering "true" to the second question independently of having knowledge about the first question.
         \item $P'(q_1|Q_2) = P(q_1)$, the belief of answering "true" to the first question independently of having knowledge about the second question.
    \end{itemize}
\end{itemize}

\begin{figure}[!h]
    \centering
    \includegraphics[scale=0.4]{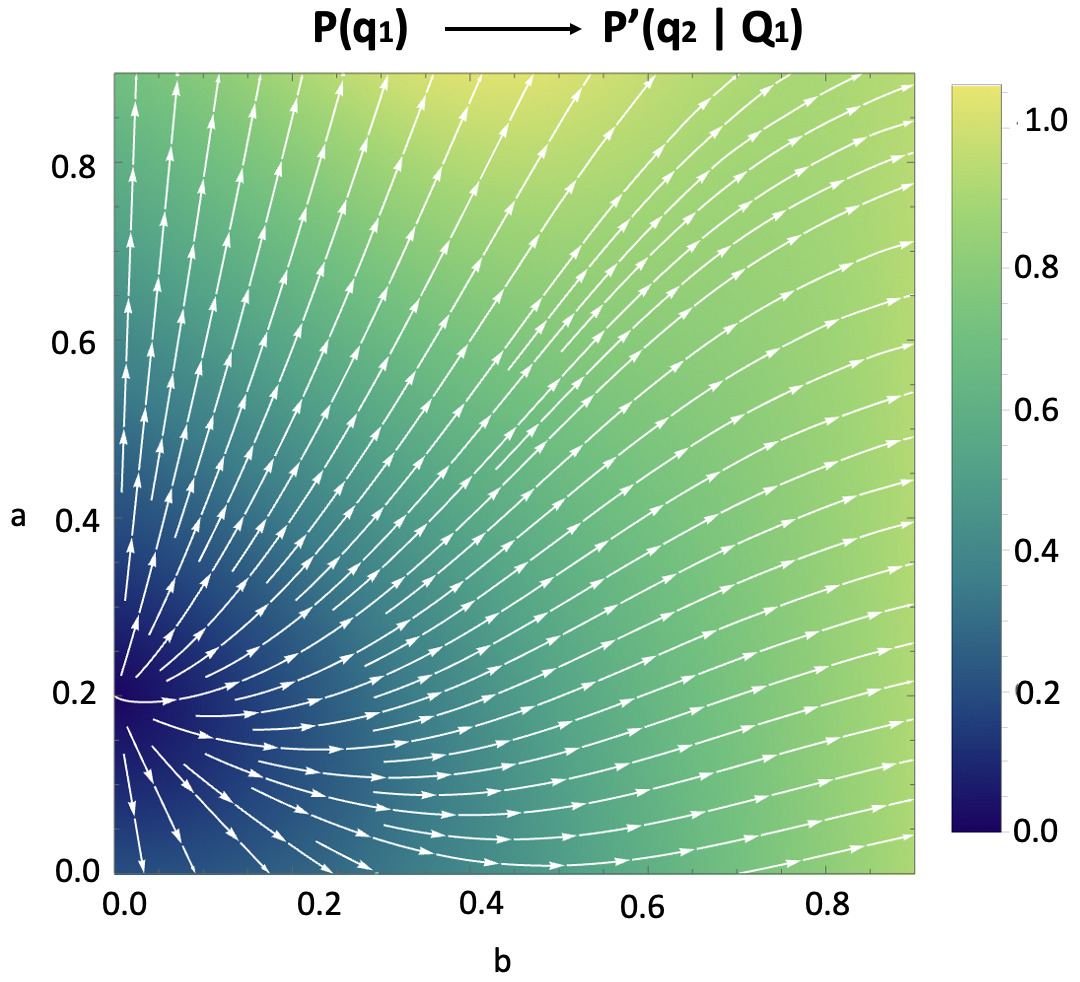}
    \caption{Order effect on belief update from answering "true" to the second question having knowledge about the first question ( Eq.\ref{eq:order_effectQ2})($c=0.2$).}
    \label{fig:bayes_update}
\end{figure}

If the above conditions are not satisfied, then we have an order effect. Figure~\ref{fig:bayes_update} illustrate an order effect in the Bayesian update from $P(q_1)$ to $P'(q_2|Q_1)$ (Equation~\ref{eq:order_effectQ2}).

\begin{figure}[!h]
    \centering
    \includegraphics[scale=0.4]{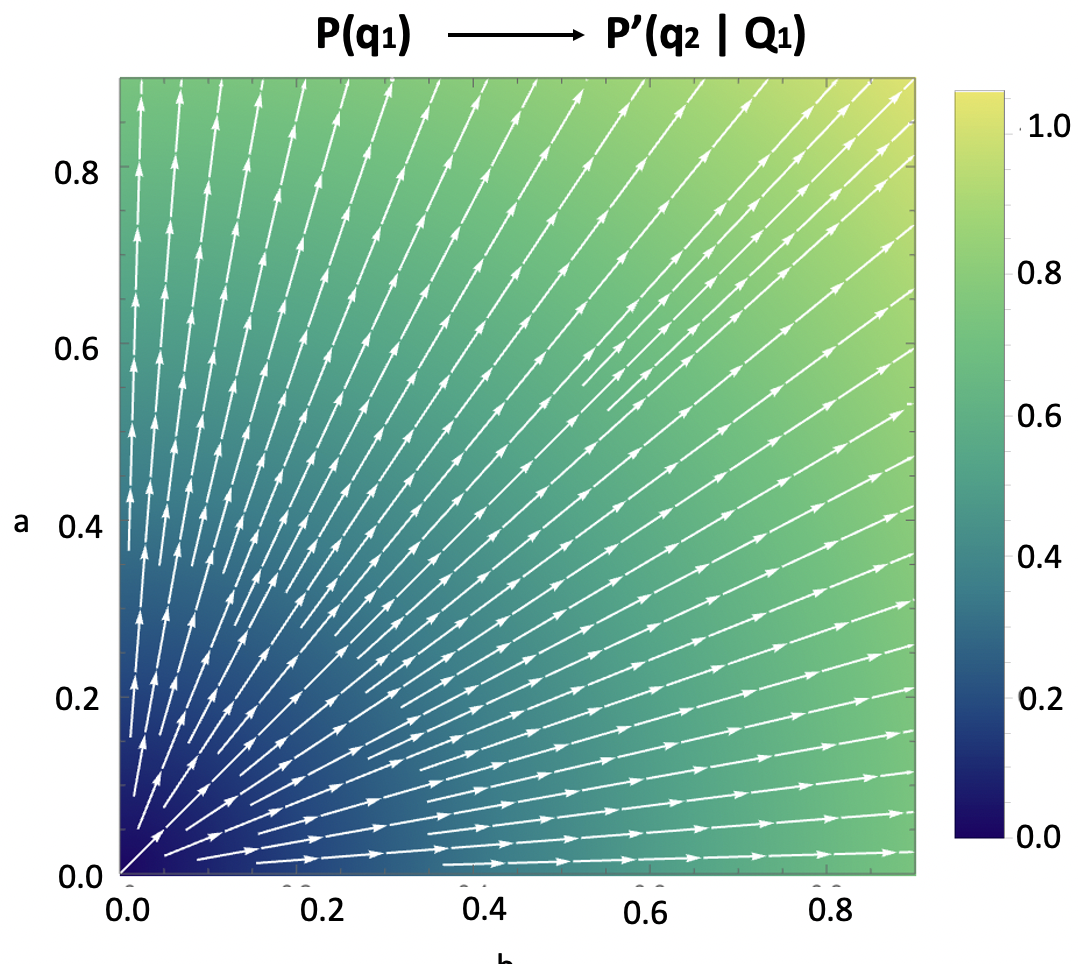}
    \caption{No order effect on belief update from answering "true" to the second question having knowledge about the first question. Setting $c=a.b$.}
    \label{fig:bayes_update_no_effect}
\end{figure}

One possibility to \textit{not have an order effect} is for $a=1/2$. This possibility is intuitive since, if we have no opinion on Clinton, it stands to reason that what a random choice will not influence how we think of Gore. The other solution, $c=a.b$, is also intuitive (and it is represented in Figure~\ref{fig:bayes_update_no_effect}): it means that our prior opinion of Gore and Clinton are statistically independent. Note that independence between questions is one of the properties that were highlighted in Eq. 3, as a necessary condition for order effects to not occur. Therefore whatever we learn about our beliefs about Clinton, they will not affect our beliefs about Gore. 

With the above example, we can understand how the update leads to the order effect. We start with a prior, say one that has Clinton at 40\% and Gore at 70\% as our belief that they are honest and trustworthy. Let us assume that, because we know Gore was Clinton's vice president, there is a correlation between our belief's about Clinton and Gore. By reflecting on Clinton and thinking, ``no, we do not think he is honest and trustworthy, and we think that the likelihood of trustworthy is just 40\%'', our prior updates and our belief about Gore changes. 

What about the QQ equality, a measure of how well the order effect matches quantum predictions. We can also compute the QQ equality for this system, and we obtain the following prediction. 
\begin{equation}
    QQ = - \frac{2(b-a)(b+a-1)(c-ab)}{(2a^2-2a+1)(2b^2-2b+1)}
\end{equation}
We can see that the QQ equality is satisfied when $a=b$, $b=1-a$, or when both Clinton and Gore are statistically independent. The latter case is not interesting since it also implies there is no order effect. Interestingly, Bayesian updates do not predict the QQ equality, but it provides conditions for which the QQ equality is satisfied. Furthermore, it allows us to try and find experimental situations where QQ is maximally violated. 

\subsection{$Q_1\rightarrow Q_2 \rightarrow Q_1$ order effect}

We now turn to a problem with quantum cognition models for order effect \citep{Khrennikov14plos}. The quantum model provided by \cite{busemeyer12question} assumes observables $\hat{A}$ and $\hat{B}$ in the Hilbert space associated with $Q_1$ and $Q_2$, such that $[\hat{A},\hat{B}]\neq 0$. If the initial cognitive state is given, then, after $Q_1$, this state collapses into a subspace of $\hat{A}$ corresponding to the eigenvalue of the response. Subsequent observation of $Q_2$ leads to another collapse into the eigenvectors of $\hat{B}$. However, this second collapse presents a problem. Since $\hat{A}$ and $\hat{B}$ do not commute, it follows that asking $Q_1$ twice yields different results if $Q_2$ is asked in between them. In other words, someone's response to $Q_1$ may change when an intermediate question is asked. 

In their paper, \cite{Khrennikov14plos} discussed this problem, claiming to be a significant issue with the quantum model. They reasoned that if one asks $Q_1$ twice, people will answer it the same. This repeatability is known as the replicability effect. In a recent article, \cite{busemeyer2017there} showed that, indeed, replicability is not necessarily true in a $Q_1\rightarrow Q_2 \rightarrow Q_1$ experiment, arguing that this supports the quantum approach. 

However, does this experiment represent a quantum effect?
%However, is this, once again, a quantum effect? 
From our Bayesian model, we can see that each time a question is asked, the prior is updated. Thus, the probability of answering $Q_1$ the second time is not the same, and  %Furthermore, it is still a probabilistic process (unless we start with zero or one priors). 
there is an
%However, we see here an 
intuitive dynamics at play for each consecutive question. The first $Q_1$ updates the prior and leads to a posterior that changes the probabilities for $Q_2$. The answering of $Q_2$ once again updates the probabilities of the new posterior for $Q_1$. One can imagine situations where, depending on the correlations and initial probabilities, $Q_1$ before $Q_2$ can be made more or less probable than the $Q_1$ after $Q_2$. 

Does the Bayesian model provide anything fundamentally different from the quantum model, as both predict the non-replicability of a question? We believe so. Consider now the further complication to the $Q_1\rightarrow Q_2 \rightarrow Q_1$ experiment: $Q_1\rightarrow Q_2 \rightarrow Q_1 \rightarrow Q_2 \rightarrow Q_1 $. The quantum model predicts that the third $Q_1$ will have the same probability as the second one. The Bayesian model, on the other hand, predicts that the probabilities will be updated every time. This change in probabilities is intuitive. The more one reconstructs an answer, the more likely we will be to repeat it.  
%In other words, the more we say something, the more it will become certain. 
This convergence to the stability of answers is not part of the quantum model, and it is a prediction of our proposed Bayesian model.  

\section{Conclusions}\label{sec:conclusions}

Order effect has been studied extensively in the literature. It is the most striking example of an experimental outcome that fits well with quantum models while challenging traditional cognitive models. Furthermore, quantum models predict some constraints for order effects, most notably the $QQ$ equality. 

In this paper, we examined the order effect as a Bayesian update phenomenon. We first showed that, by thinking about questions in terms of a Bayesian Network, one could realize that some traditional models, such as those discussed by \cite{trueblood2011quantum}, fail because they do not account for temporal processes. Notably, they do not incorporate changes in the probability distribution during a Bayesian update. To address this challenge, we constructed an explicit Bayesian update model where each question can be thought of as a mini-experiment where the respondent reflects on their beliefs. For this model, we showed that order effects appear, and they have a simple cognitive explanation for their existence: the respondent's prior belief that two questions are correlated. 

Similar to the quantum order effect model, our Bayesian model allows us to make several predictions. First, we see certain conditions on the prior that limit the existence of order effects. For example, if $Q_1$ and $Q_2$ are statistically independent, one should not expect order effects. These conditions allow us to think about possible experiments where order effects will be strong or non-existent. Second, we show that, for our model, the $QQ$ equality is not necessarily satisfied. This does not mean that $QQ$ will not be satisfied for a particular experiment. It only means that $QQ$ requires further assumptions (i.e., certain symmetry conditions). Third, like the quantum model, the Bayesian model does not predict the exact replicability of questions. However, contrary to the quantum model, the more questions are asked, the more the posterior will converge, stabilizing the answer. Finally, the Bayesian model has the advantage of having fewer parameters than the quantum model. For the simplest order-effect quantum model, a minimum of four parameters is necessary. For our Bayesian model, three parameters describe the joint probability distributions entirely. Whether the Bayesian model is better than the quantum model is a question to be settled empirically, which we are planning to address as future work.

\section{Acknowledgements}

This work was supported by  Centre for Data Science First Byte Funding Program at Queensland University of Technology (QUT) and by QUT's Women in Research Grant Scheme.

\setlength{\bibleftmargin}{.125in}
\setlength{\bibindent}{-\bibleftmargin}
\bibliographystyle{apacite}
%\bibliography{references}

\end{document}